\documentclass[10pt,twocolumn,letterpaper]{article}

\usepackage{iccv}
\usepackage{times}
\usepackage{epsfig}
\usepackage{graphicx}
\usepackage{amsmath}
\usepackage{amssymb}

\usepackage{booktabs}
\usepackage{multirow}
\usepackage{subfigure}
\usepackage{color}

\usepackage[breaklinks=true,bookmarks=false]{hyperref}

\iccvfinalcopy 

\def\iccvPaperID{****} 

\ificcvfinal\fi

\begin{document}

\title{Universal-Prototype Enhancing for Few-Shot Object Detection}

\author{Aming Wu$^{1}$ \quad Yahong Han$^{2,3,4}$\thanks{Corresponding author} \quad Linchao Zhu$^{5}$ \quad Yi Yang$^{5}$\\
$^{1}$School of Electronic Engineering, Xidian University, Xi'an, China\\
$^{2}$College of Intelligence and Computing, Tianjin University, Tianjin, China\\
$^{3}$Tianjin Key Lab of Machine Learning, Tianjin University, Tianjin, China\\
$^{4}$Peng Cheng Laboratory, Shenzhen, China \quad
$^{5}$ReLER Lab, AAII, University of Technology Sydney\\
amwu@xidian.edu.cn, \quad yahong@tju.edu.cn, \quad \{Linchao.Zhu, yi.yang\}@uts.edu.au
}

\maketitle
\ificcvfinal\thispagestyle{empty}\fi

\begin{abstract}
   Few-shot object detection (FSOD) aims to strengthen the performance of novel object detection with few labeled samples. To alleviate the constraint of few samples, enhancing the generalization ability of learned features for novel objects plays a key role. Thus, the feature learning process of FSOD should focus more on intrinsical object characteristics, which are invariant under different visual changes and therefore are helpful for feature generalization. Unlike previous attempts of the meta-learning paradigm, in this paper, we explore how to enhance object features with intrinsical characteristics that are universal across different object categories. We propose a new prototype, namely universal prototype, that is learned from all object categories. Besides the advantage of characterizing invariant characteristics, the universal prototypes alleviate the impact of unbalanced object categories. After enhancing object features with the universal prototypes, we impose a consistency loss to maximize the agreement between the enhanced features and the original ones, which is beneficial for learning invariant object characteristics. Thus, we develop a new framework of few-shot object detection with universal prototypes (${FSOD}^{up}$) that owns the merit of feature generalization towards novel objects. Experimental results on PASCAL VOC and MS COCO show the effectiveness of ${FSOD}^{up}$. Particularly, for the 1-shot case of VOC Split2, ${FSOD}^{up}$ outperforms the baseline by 6.8\% in terms of mAP. 
\end{abstract}

\section{Introduction}

\begin{figure}
\centering
\includegraphics[width=1.0\linewidth]{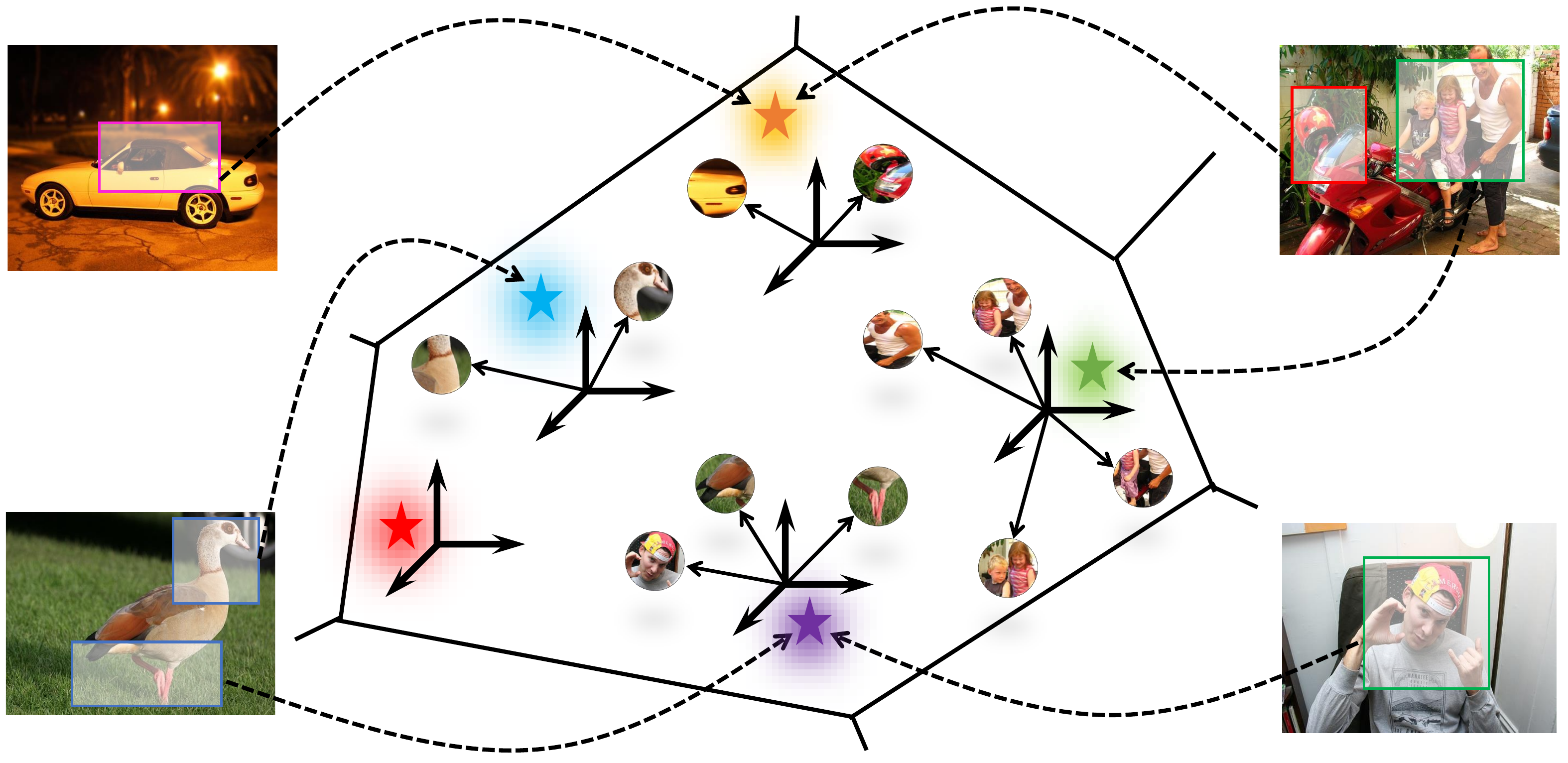}
\caption{Universal prototypes (colorful stars) are learned from all object categories, which are not specific to certain object categories. Universal prototypes capture different intrinsical object characteristics via latent projection, e.g., the prototype \textcolor[rgb]{1.00,0.5,0.00}{$\bigstar$} incorporates object characteristics of `car' and `motorbike'.}
\label{fig1}
\vspace{-0.1in}
\end{figure}

Recently, owing to the success of deep learning, great progress has been made on object detection \cite{ren2015faster,girshick2015fast,he2017mask,girshick2014rich}. However, the outstanding performance \cite{redmon2017yolo9000,liu2016ssd,carion2020end,lin2017feature} depends on abundant annotated objects in training images for each category. As a challenging task, few-shot object detection (FSOD) \cite{kang2019few,wang2019meta} mainly aims to improve the detection performance for novel objects that belong to certain categories but appear rarely in the annotated training images.

The main challenge of FSOD lies in how to learn generalized object features from both abundant samples in base categories and few samples in novel categories, which can simultaneously describe invariant object characteristics and alleviate the impact of unbalanced categories. Recently, meta-learning strategy \cite{snell2017prototypical,tian2020rethinking,finn2017model} has been utilized in \cite{yan2019meta,xiao2020few,wang2019meta,fan2020few} to adapt representation ability from base object categories to novel categories. However, the weak performance compared to basic fine-tuning methods \cite{wang2020frustratingly,wu2020multi,chen2019a,dhillon2020a} shows the meta-learning technique fails to improve the generalization ability of object feature learning.

One possible reason is that the adaptation process in meta-learning mechanism could not capture the invariant characteristics across categories sufficiently. The invariance, i.e., invariant under different visual changes like textual variances or environmental noises, is always associated with the intrinsical object characteristics. As demonstrated in \cite{lyle2019analysis}, the models that could extract invariant representations often generalize better than their non-invariant counterparts. Therefore, in this paper, we explore how to enhance the generalization ability of object feature learning with the invariant object characteristics.

We devise universal prototypes (as shown in Fig. \ref{fig1}) to learn the invariant object characteristics. Different from the prototypes that are separately learned from each category \cite{snell2017prototypical,liu2019prototype,wang2019panet}, the proposed universal prototypes are learned from all object categories. The benefits are two-fold. On the one hand, prototypes from all categories capture rich information not only from different object categories but also from contexts of images. On the other hand, the universal prototypes reduce the impact of data-imbalance across different categories. Moreover, via fine-tuning, the universal prototypes can be effectively adapted to data-scarce novel categories. To this end, we develop a new framework of few-shot object detection with universal prototypes (${FSOD}^{up}$). Particularly, we utilize a soft-attention of the learned universal prototypes to enhance the object features. Such a universal-prototype enhancement (i.e., each element of the enhanced features is a combination of prototypes) aims to simultaneously improve invariance and retain the semantic information of original object features. Here we employ a consistency loss to enable the maximum agreement between the enhanced and original object features. During training, we first train the model on data-abundant base categories. Then, the model is fine-tuned on a re-constructed training set that contains a small number of balanced training samples from both base and novel object categories. Experimental results on two benchmarks and extensive visualization analyses demonstrate the effectiveness of the proposed method. Our code will be available at \url{https://github.com/AmingWu/UP-FSOD}.

The contributions are summarized as follows:

(1) Towards FSOD, we devise a dedicated prototype and a new framework with universal-prototype enhancenment.

(2) We successfully demonstrate that, after fine-tuning with universal-prototype enhanced features, object detectors effectively adapt to novel categories.

(3) We obtain new performance on PASCAL VOC \cite{everingham2010pascal,everingham2015pascal} and MS COCO \cite{lin2014microsoft}. Enhancing invariance and generalization with the learned universal prototypes is empirically verified. Moreover, extensive visualization analyses also show that universal prototypes are capable of enhancing object characteristics, which is beneficial for FSOD.

\section{Related Work}

\textbf{Few-shot image classification.} Few-shot image classification \cite{vinyals2016matching,ravi2017optimization,sung2018learning,hariharan2017low,gidaris2018dynamic} targets to recognize novel categories with only few samples in each category. Meta-learning is a widely used method to solve few-shot classification \cite{lu2020learning}, which aims to leverage task-level meta knowledge to help the model adapt to new tasks with few labeled samples. Vinyals et al. \cite{vinyals2016matching} and Snell et al. \cite{snell2017prototypical} employed the meta-learning policy to learn the similarity metric that could be transferrable across different tasks. Particularly, based on the policy of meta-learning, prototypical network \cite{snell2017prototypical} is proposed to take the center of congener support samples' embeddings as the prototype of this category. The classification can be performed by computing distances between the representations of samples and prototype of each category. However, when the data is unbalanced or scarce, the learned prototypes could not represent the information of each category accurately, which affects the classification performance. Besides, during meta-learning, Gidaris et al. \cite{gidaris2018dynamic} and Wang et al. \cite{wang2019tafe} introduced new parameters to promote the adaptation to novel tasks. However, these meta-learning methods for few-shot image classification could not be directly applied to object detection that requires localizing and recognizing objects.

\begin{figure*}
\centering
\includegraphics[width=1.0\linewidth]{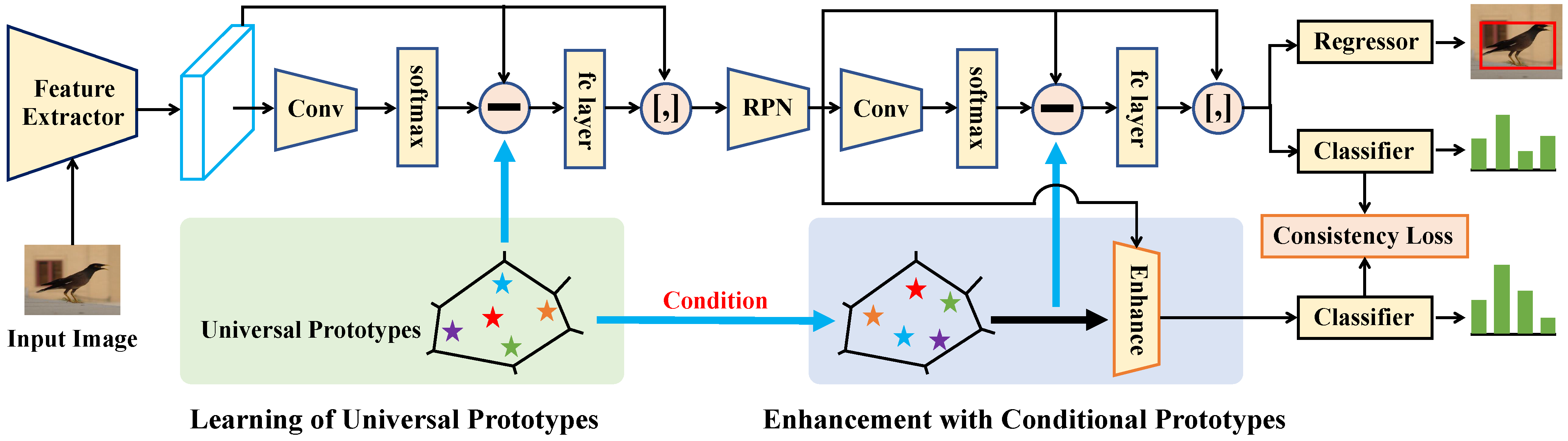}
\caption{The architecture of few-shot object detection with universal-prototype enhancement. `Conv' and `fc layer' separately indicate convolution and fully-connected layer. The colorful stars are the learned universal prototypes. `$\ominus$' and `[,]' denote the residual operation and concatenation operation, respectively. We focus on improving the generalization of detectors via learning invariant object characteristics. Firstly, universal prototypes are learned from all object categories. With the output of RPN (Region Proposal Network), we obtain the conditional prototypes via a conditional transformation of universal prototypes. Next, the enhanced object features are calculated based on conditional prototypes. Finally, a consistency loss is computed between the enhanced and original features.}
\label{fig2}
\vspace{-0.1in}
\end{figure*}

\textbf{Few-shot object detection.} Most existing methods employ meta-learning \cite{fan2020few,Karlinsky_2019_CVPR} or fine-tuning \cite{yang2020context,wu2020multi} strategies to solve FSOD. Specifically, Wang et al. \cite{wang2019meta} developed a meta-learning based framework to leverage meta-level knowledge from data-abundant base categories to learn a detector for novel categories. Yan et al. \cite{yan2019meta} further extended Faster R-CNN \cite{ren2015faster} by performing meta-learning over RoI (Region-of-Interest) features. However, the weak performance compared to basic fine-tuning methods shows meta-learning based methods fail to improve the generalization ability of object detectors. For the method of fine-tuning and the model pre-trained on the base categories, Wang et al. \cite{wang2020frustratingly} employed a two-stage fine-tuning process, i.e., fine-turning the last layers of the detector and freezing the other parameters of the detector, to make the object predictor adapt to novel categories. Wu et al. \cite{wu2020multi} proposed a method of multi-scale positive sample refinement to handle the problem of scale variations in object detection, which is similar to data augmentation \cite{zoph2019learning}.

Different from previous methods for FSOD, in this paper, we propose to learn universal prototypes from all object categories. And we develop a new framework of FSOD with universal-prototype enhancement. Experimental results and visualization analysis demonstrate the effectiveness of universal-prototype enhancement.

\section{FSOD with Universal Prototypes}

In this paper, we follow the same FSOD settings introduced in Kang et al. \cite{kang2019few}. Annotated detection data are divided into a set of base categories that have abundant instances and a set of novel categories that have only few (usually less than 30) instances per category. The main purpose is to improve the generalization ability of detectors.

\subsection{Learning of Universal Prototypes}

Recently, many methods \cite{snell2017prototypical,liu2019prototype,wang2019panet} construct a prototype for each category to solve few-shot image classification. Though prototypes reflecting category information have been demonstrated to be effective for image classification, they could not be applied to FSOD. The reason may be that these category-specific prototypes represent image-level information and fail to capture object characteristics that are helpful for localizing and recognizing objects. Different from category-specific prototypes, based on all object categories, we attempt to learn universal prototypes that are beneficial for capturing intrinsical object characteristics that are invariant under different visual changes.

Concretely, the left part of Fig. \ref{fig2} shows the learning process of universal prototypes. We adopt widely used Faster R-CNN \cite{ren2015faster}, a two-stage object detector, as the base detection model. Given an input image, we first employ the feature extractor, e.g., ResNet \cite{he2016deep}, to extract corresponding features $F \in \mathbb{R}^{w \times h \times m}$, where $w$, $h$, and $m$ separately denote width, height, and the number of channels. Then, the universal prototypes are defined as $C = \{c_{i} \in \mathbb{R}^{m}, i=1,...,D\}$. Next, based on the prototypical set $C$, we calculate descriptors that represent image-level information.
\begin{equation}\label{eq1}
\begin{split}
& \mathcal{I} = W_{g} * F + b_{g}, \\
& \mathcal{V}_{i} = \sum_{j=1}^{wh} \frac{e^{\mathcal{I}_{j,i}}}{\sum_{i=1}^{D} e^{\mathcal{I}_{j,i}}} (F_{j} - c_{i}),
\end{split}
\end{equation}
where $W_{g} \in \mathbb{R}^{3 \times 3 \times m \times D}$ and $b_{g} \in \mathbb{R}^{D}$ are convolutional parameters. $\mathcal{V} \in \mathbb{R}^{D \times m}$ represents the output descriptors. `$F_{j} - c_{i}$' indicates the residual operation, by which the visual features can be assigned to the corresponding prototype. Finally, we take the concatenated result of $F$ and $\mathcal{V}$ as the input of the RPN module.
\begin{equation}\label{eq2}
P = {\rm RPN}(\Psi([F, ~\mathcal{V}_{r}W_{p} + b_{p}])),
\end{equation}
where $\mathcal{V}_{r} \in \mathbb{R}^{1 \times Dm}$ is the reshaped result of $\mathcal{V}$. Meanwhile, $W_{p} \in \mathbb{R}^{Dm \times m}$ and $b_{p} \in \mathbb{R}^{m}$ are parameters of the fully-connected layer. `[,]' is the concatenation operation. By the concatenation operation, the descriptors $\mathcal{V}$ can be fused into the original features $F$, which enhances the representation ability of $F$. $\Psi$ consists of two convolutional layers with ReLU activation and is used to transform the concatenated result. Finally, $P \in \mathbb{R}^{n \times s \times s \times m}$ is the output of RPN with RoI Pooling \cite{ren2015faster,he2017mask}, where $n$ and $s$ separately indicate the number of proposals and the size of proposals. The feature dimension of $P$ is the same as $F$.

\subsection{Enhancement of Object Features}

As shown in the right part of Fig. \ref{fig2}, we first compute conditional prototypes based on the universal prototypes $C$. Then, we conduct enhancement of object features with the conditional prototypes.

\subsubsection{The Computation of Conditional Prototypes}

Since the computation of Eq. \eqref{eq1} is based on the extracted features that represent the whole input image, the universal prototypes $C$ mainly reflect image-level information. Here, the image-level information includes object-level information and other associated information about image content. Whereas, after RPN, the proposal features $P$ mainly contain object-level information. The directly using of universal prototypes $C$ may not accurately represent object-level information. Thus, we make an affine transformation to promote $C$ to move towards the space of object-level features.
\begin{equation}\label{eq3}
\mathcal{A} = \alpha \odot C + \beta,
\end{equation}
where $\alpha \in \mathbb{R}^{D \times 1}$ and $\beta \in \mathbb{R}^{D \times 1}$ are the transformed parameters. $\odot$ is element-wise product. Finally, $\mathcal{A} \in \mathbb{R}^{D \times m}$ represents the conditional prototypes. Next, we employ the same processes as Eq. \eqref{eq1} to generate object-level descriptors. The processes are shown as follows:
\begin{equation}\label{eq4}
\begin{split}
& E = W_{c} * P + b_{c}, \\
& O_{k,i} = \sum_{j=1}^{s^{2}} \frac{e^{E_{k,j,i}}}{\sum_{i=1}^{D} e^{E_{k,j,i}}} (P_{k,j} - a_{i}),
\end{split}
\end{equation}
where $k=1, \cdots, n$. $W_{c} \in \mathbb{R}^{3 \times 3 \times m \times D}$ and $b_{c} \in \mathbb{R}^{D}$ are convolutional parameters. $a_{i} \in \mathbb{R}^{1 \times m}$ is the $i$-th conditional prototype of $\mathcal{A}$. $O \in \mathbb{R}^{n \times D \times m}$ indicates the output descriptors. Finally, we take the concatenated result of $P$ and $O$ as the input of the classifier.
\begin{equation}\label{eq5}
y = {\rm Clf}([\Psi_{c}(P), ~O_{r}W_{r} + b_{r}]),
\end{equation}
where $O_{r} \in \mathbb{R}^{n \times Dm}$ is the reshaped result of $O$. ${\rm Clf}$ denotes the classifier. Meanwhile, $W_{r} \in \mathbb{R}^{Dm \times 2m}$ and $b_{r} \in \mathbb{R}^{2m}$ are parameters of the fully-connected layer. $\Psi_{c}$ consists of two fully-connected layers and outputs a matrix with the dimension $n \times 2m$. Finally, $y$ is the predicted probability. In the experiment, we find employing the descriptors $O$ generated based on the conditional prototypes improves the performance of FSOD, which shows the effectiveness of conditional prototypes.

\subsubsection{Enhancement with Conditional Prototypes}

\begin{figure}[ht]
\centering
\includegraphics[width=1.0\linewidth]{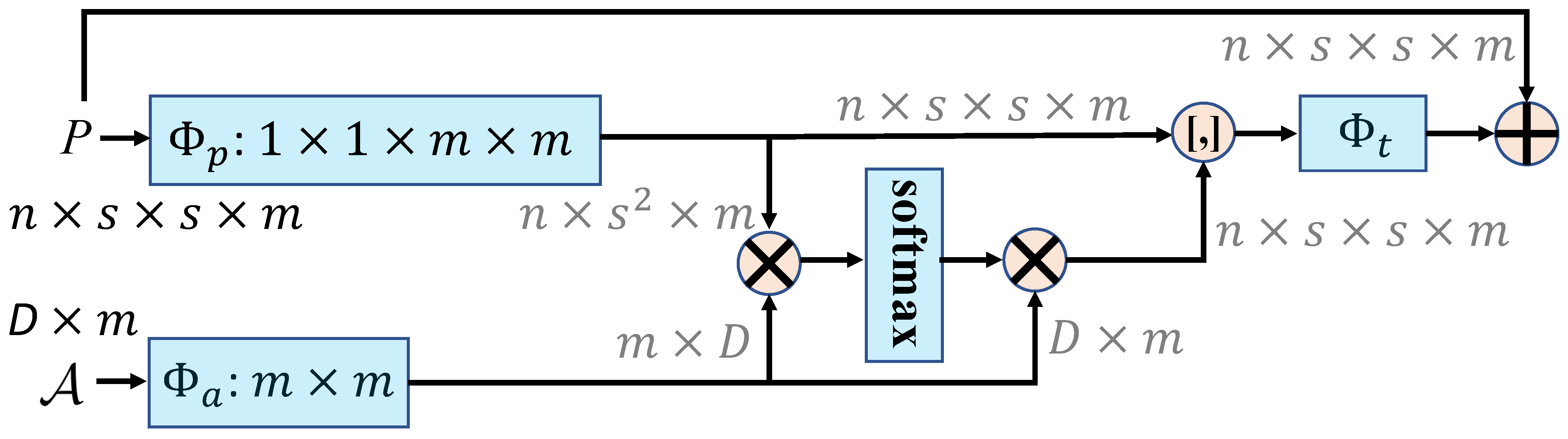}
\caption{Enhancement of object features. Based on each element of RPN output $P$, we calculate the soft-attention of the conditional prototypes $\mathcal{A}$ to generate enhanced features. Each element of the enhanced features is a combination of conditional prototypes, which retains the semantic information of $P$.}
\label{fig3}
\vspace{-0.1in}
\end{figure}

In order to improve the generalization of detectors, we explore to utilize conditional prototypes to enhance object features. Specifically, Fig. \ref{fig3} shows the enhancement details. For proposal features $P \in \mathbb{R}^{n \times s \times s \times m}$ and conditional prototypes $\mathcal{A} \in \mathbb{R}^{D \times m}$, we separately employ a convolutional layer $\Phi_{p} \in \mathbb{R}^{1 \times 1 \times m \times m}$ and fully-connected layer $\Phi_{a} \in \mathbb{R}^{m \times m}$ to project $P$ and $\mathcal{A}$ into an embedding space, i.e., $e_{p} = \Phi_{p}(P)$ and $e_{a} = \Phi_{a}(\mathcal{A})$. Then, based on each element of $e_{p}$, we calculate the soft-attention of $e_{a}$ to obtain enhancement of object features.
\begin{equation}\label{eq6}
\begin{split}
& \lambda_{k} = {\rm softmax}(e_{p,k}e_{a}^{T}), \\
& Enh_{k} = {\rm ReLU}(\Phi_{t}([e_{p,k},~\lambda_{k}e_{a}]) + P_{k}),
\end{split}
\end{equation}
where $k = 1, \cdots, n$. $e_{p,k} \in \mathbb{R}^{s^{2} \times m}$ indicates the $k$-th component of $e_{p}$. $\lambda_{k} \in \mathbb{R}^{s^{2} \times D}$ denotes attention weights. $\Phi_{t}$ consists of two convolutional layers with ReLU activation. And the output dimension of $\Phi_{t}$ is $m$. $P_{k} \in \mathbb{R}^{s \times s \times m}$ is the $k$-th component of $P$. Finally, $Enh \in \mathbb{R}^{n \times s \times s \times m}$ is the enhanced object features, which fuses the information of conditional prototypes and is helpful for improving the generalization on novel objects. Next, $Enh$ is taken as the input of the classifier to output the predicted probability.
\begin{equation}\label{eq7}
y_{enh} = {\rm Clf}([\Psi_{c}(Enh), ~\Psi_{c}(P)]),
\end{equation}
where $y_{enh}$ is the predicted probability. Besides, Eq. \eqref{eq5} and Eq. \eqref{eq7} share the same classifier. In the experiment, we find the enhanced operations (Eq. \eqref{eq6} and \eqref{eq7}) are beneficial for FSOD, which further indicates the learned prototypes contain object-level information.

\begin{figure*}[ht]
\centering
\includegraphics[width=1.0\linewidth]{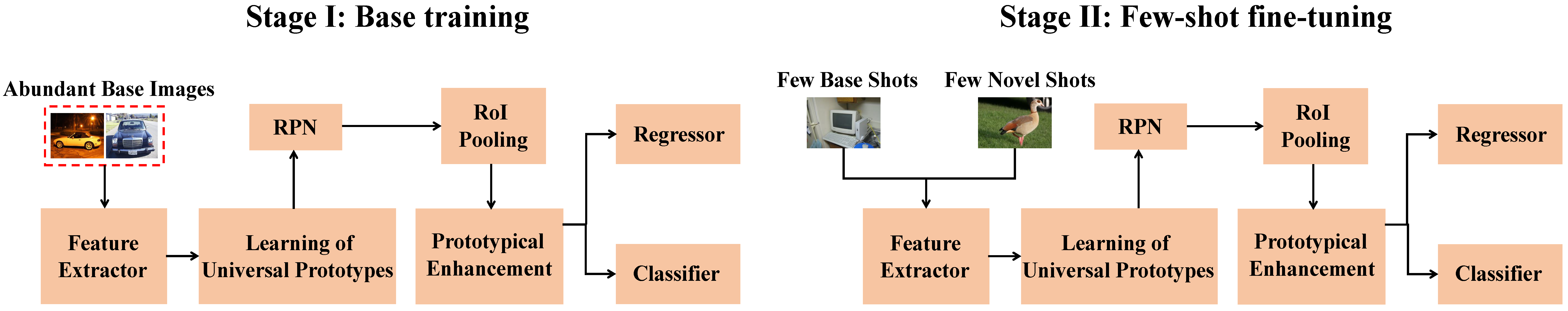}
\caption{Illustration of two-stage fine-tuning approach for ${FSOD}^{up}$. In the base training stage, the entire detector, including the feature extractor, the module for learning of universal prototypes, and the module for enhancement based on conditional prototypes, are jointly trained on the data-abundant base categories. In the few-shot fine-tuning stage, the entire detector is fine-tuned on a balanced training set consisting of both the few base and novel categories.}
\label{fig4}
\vspace{-0.1in}
\end{figure*}

\subsection{Two-stage Fine-tuning Approach}

Many semi-supervised learning methods \cite{berthelot2019mixmatch,berthelot2019remixmatch} rely on a consistency loss to enforce that the model output remains unchanged when the input is perturbed. Inspired by this idea, to learn invariant object characteristics, we compute the consistency loss between the prediction $y$ from original features (see Eq. \eqref{eq5}) and the prediction $y_{enh}$ from enhanced features. Particularly, the KL-Divergence loss is employed to enforce consistent predictions, i.e., $\mathcal{L}_{\rm con} = \mathcal{H}(y, ~y_{enh})$. The joint training loss is defined as follows:
\begin{equation}\label{eq8}
\mathcal{L} = \mathcal{L}_{\rm rpn} + \mathcal{L}_{\rm cls} + \mathcal{L}_{\rm loc} + \gamma\mathcal{L}_{\rm con},
\end{equation}
where $\mathcal{L}_{\rm rpn}$ is the loss of the RPN to distinguish foreground from background and refine bounding-box anchors. $\mathcal{L}_{\rm cls}$ and $\mathcal{L}_{\rm loc}$ separately indicate classification loss and box regression loss. And $\gamma$ is a hyper-parameter.

During training, we employ a two-stage fine-tuning approach (as shown in Fig. \ref{fig4}) to optimize ${FSOD}^{up}$ model. Concretely, in the base training stage, we employ the joint loss $\mathcal{L}$ to optimize the entire model based on the data-abundant base classes. After the base training stage, only the last fully-connected layer (for classification) of the detection head is replaced. The new classification layer is randomly initialized. Besides, during few-shot fine-tuning stage, different from the work \cite{wang2020frustratingly}, none of the network layers is frozen. And we still employ the loss $\mathcal{L}$ to fine-tune the entire model based on a balanced training set consisting of both the few base and novel categories.

\subsection{Discussion}

In this section, we further discuss universal prototypes for few-shot object detection.

Though prototypes have been demonstrated to be effective for few-shot image classification \cite{snell2017prototypical,vinyals2016matching}, it is unclear how to build prototypes for FSOD \cite{kang2019few}.
(1) If we follow few-shot image classification and construct prototypes for each category, the computational costs increase for the case of a large number of object categories. Meanwhile, due to the unbalanced object categories, the constructed prototypes may not accurately reflect category information. (2) Related to the above, detectors for certain object category can be affected by co-appearing objects in one image, and thus the quality of the constructed prototype for such category may be burdened. (3) More importantly, since the number of object categories in the stage of the base training is different from that of the few-shot fine-tuning, constructing a prototype for each object category makes it impossible to align the prototypes between the base training and the few-shot fine-tuning. That is to say, the prototypes pre-trained on base categories cannot be directly utilized in the fine-tuning stage. Therefore, for fine-tuning based methods, it is difficult to build a prototype for each category.

To solve FSOD, we propose to learn universal prototypes from all object categories. The universal prototypes are not specific to certain object categories and can be effectively adapted to novel categories via fine-tuning. In the experiments, we find that the universal prototypes are helpful for characterizing the regional information of different object categories. Meanwhile, with the help of universal-prototype enhancement, the performance of few-shot detection can be significantly improved.

\section{Experiments}

We first evaluate our method on PASCAL VOC \cite{everingham2010pascal,everingham2015pascal} and MS COCO \cite{lin2014microsoft}. For a fair comparison, we use the settings in \cite{kang2019few,yan2019meta} to construct few-shot detection datasets. Concretely, for PASCAL VOC, the 20 classes are randomly divided into 5 novel classes and 15 base classes. Here, we follow the work \cite{kang2019few} to use the same three class splits, where only $K$ object instances are available for each novel category and $K$ is set to 1, 2, 3, 5, 10. For MS COCO, the 20 categories overlapped with PASCAL VOC are used as novel categories with $K$ = 10, 30. And the remaining 60 categories are taken as the base categories.

\textbf{Implementation Details.} Faster R-CNN \cite{ren2015faster} is used as the base detector. Our backbone is Resnet-101 \cite{he2016deep} with the RoI Align \cite{he2017mask} layer. We use the weights pre-trained on ImageNet \cite{russakovsky2015imagenet} in initialization. For FSOD, the number of universal prototypes (see Eq. \eqref{eq1}) is set to 24. All these prototypes are randomly initialized. Next, the model is trained with a batchsize of 2 on 2 GPUs, 1 image per GPU. Meanwhile, to alleviate the impact of the scale issue, we employ the positive sample refinement \cite{wu2020multi}. The hyper-parameter $\gamma$ (see Eq. \eqref{eq8}) is set to 1.0. All models are trained using SGD optimizer with a momentum of 0.9 and a weight decay of 0.0001. Finally, during inference, we take the output $y$ of Eq. \eqref{eq5} as the classification result.

\subsection{Performance Analysis of Few-Shot Detection}

We compare ${FSOD}^{up}$ with two baseline methods, i.e., TFA \cite{wang2020frustratingly} and MPSR \cite{wu2020multi}. These two approaches all use the two-stage fine-tuning method to solve FSOD.

\textbf{Results on PASCAL VOC.} Table \ref{table1} shows the results of PASCAL VOC. As the number of novel categories decreases, the performance degrades significantly. This indicates that addressing the few-shot problem is crucial to improve the generalization of detectors. We can see that the proposed ${FSOD}^{up}$ method consistently outperforms the two baseline methods. This shows that employing universal-prototype enhancement is helpful for learning invariant object characteristics and thus improves performance. Meanwhile, this also indicates that focusing on invariance plays a key role in solving FSOD.

\begin{table*}[!ht]
\centering
\small
\setlength{\tabcolsep}{0.6em}
\begin{tabular}{l|ccccc|ccccc|ccccc}
\toprule
\multicolumn{1}{c|}{} & \multicolumn{5}{c|}{Novel Set 1} & \multicolumn{5}{c|}{Novel Set 2} & \multicolumn{5}{c}{Novel Set 3} \\ \midrule
Method / Shot         & 1     & 2     & 3    & 5    & 10   & 1     & 2     & 3    & 5    & 10   & 1     & 2     & 3    & 5    & 10   \\ \midrule
Meta R-CNN \cite{yan2019meta}   & 19.9   & 25.5   & 35.0  & 45.7  & 51.5  & 10.4   & 19.4   & 29.6  & 34.8  & 45.4  & 14.3   & 18.2   & 27.5  & 41.2  & 48.1 \\
RepMet \cite{Karlinsky_2019_CVPR}    & 26.1 & 32.9 & 34.4 & 38.6 & 41.3 & 17.2  & 22.1 & 23.4 & 28.3 & 35.8 & 27.5  & 31.1 & 31.5 & 34.4 & 37.2 \\
FSOD-VE \cite{xiao2020few}  & 24.2 & 35.3 & 42.2 & 49.1 & 57.4 & 21.6  & 24.6  & 31.9 & 37.0 & 45.7 & 21.2 & 30.0 & 37.2 & 43.8 & 49.6 \\ \midrule
TFA w/fc \cite{wang2020frustratingly} & 36.8 & 29.1 & 43.6 & \textbf{55.7} & 57.0 & 18.2 & 29.0 & 33.4 & 35.5 & 39.0 & 27.7 & 33.6 & 42.5 & 48.7 & 50.2 \\
TFA w/cos \cite{wang2020frustratingly} & 39.8 & 36.1 & 44.7 & \textbf{55.7} & 56.0 & 23.5 & 26.9 & 34.1 & 35.1 & 39.1 & 30.8 & 34.8 & 42.8 & 49.5 & 49.8 \\
${\rm TFA}^{\sharp}$ w/fc \cite{xiao2020few,wang2020frustratingly} & 22.9 & 34.5 & 40.4 & 46.7 & 52.0 & 16.9 & 26.4 & 30.5 & 34.6 & 39.7 & 15.7 & 27.2 & 34.7 & 40.8 & 44.6 \\
${\rm TFA}^{\sharp}$ w/cos \cite{xiao2020few,wang2020frustratingly} & 25.3 & 36.4 & 42.1 & 47.9 & 52.8 & 18.3 & 27.5 & 30.9 & 34.1 & 39.5 & 17.9 & 27.2 & 34.3 & 40.8 & 45.6 \\
${\rm MPSR}^{\sharp}$ \cite{wu2020multi} & 40.7 & 41.2 & 48.9 & 53.6 & 60.3 & 24.4 & 29.3 & 39.2 & 39.9 & 47.8 & 32.9 & 34.4 & 42.3 & 48.0 & 49.2 \\ \midrule
Ours (${FSOD}^{up}$)  & \textbf{43.8}  & \textbf{47.8}  & \textbf{50.3} & \textbf{55.4} & \textbf{61.7} & \textbf{31.2}  & \textbf{30.5}  & \textbf{41.2} & \textbf{42.2} & \textbf{48.3} & \textbf{35.5}  & \textbf{39.7}  & \textbf{43.9} & \textbf{50.6} & \textbf{53.5} \\ \bottomrule
\end{tabular}
\vspace{0.1in}
\caption{Few-shot detection performance (mAP (\%)) on PASCAL VOC dataset. We evaluate the performance on three different sets of novel categories. Resnet-101 \cite{he2016deep} is used as the backbone. `$\sharp$' indicates that we directly run the released code to obtain the results.
}\label{table1}
\vspace{-0.1in}
\end{table*}

\begin{figure}[ht]
\begin{center}
  \subfigure{
  \begin{minipage}[t]{0.28\linewidth}
    \includegraphics[width=1.0in,height=1.0in]{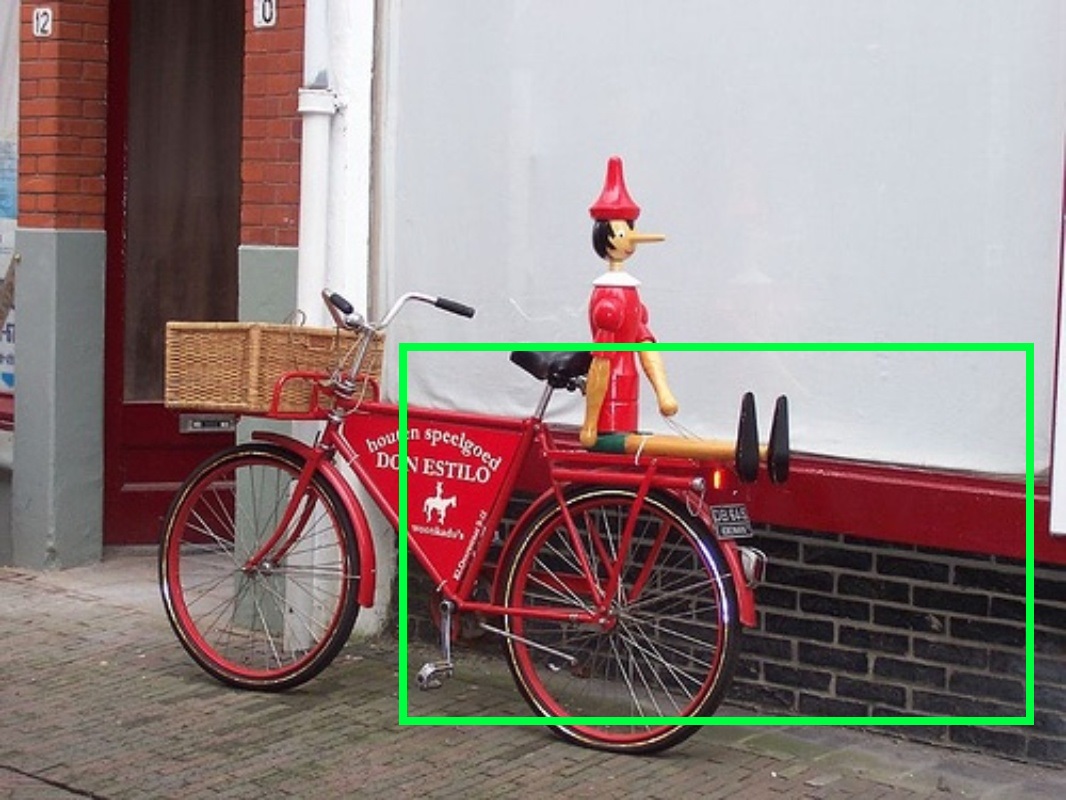}\\ \vspace{-0.1in}
    \includegraphics[width=1.0in,height=1.0in]{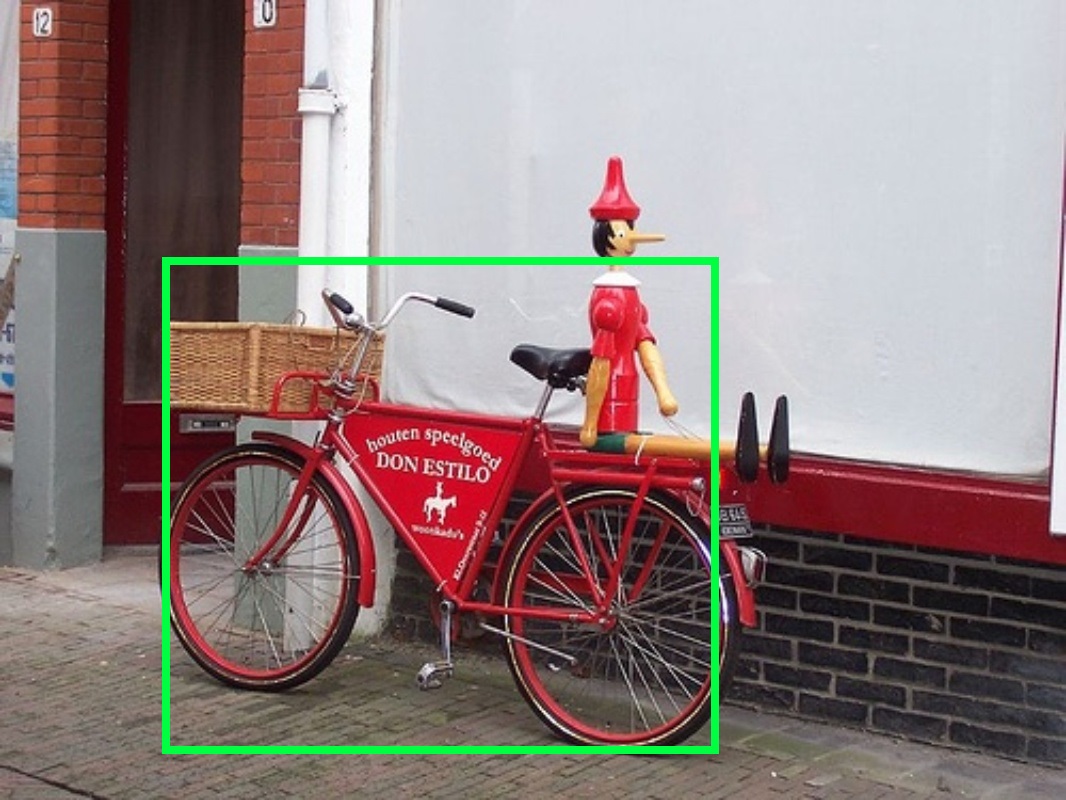}
  \end{minipage}
  }
  \hspace{0.01cm}
  \subfigure{
  \begin{minipage}[t]{0.28\linewidth}
    \includegraphics[width=1.0in,height=1.0in]{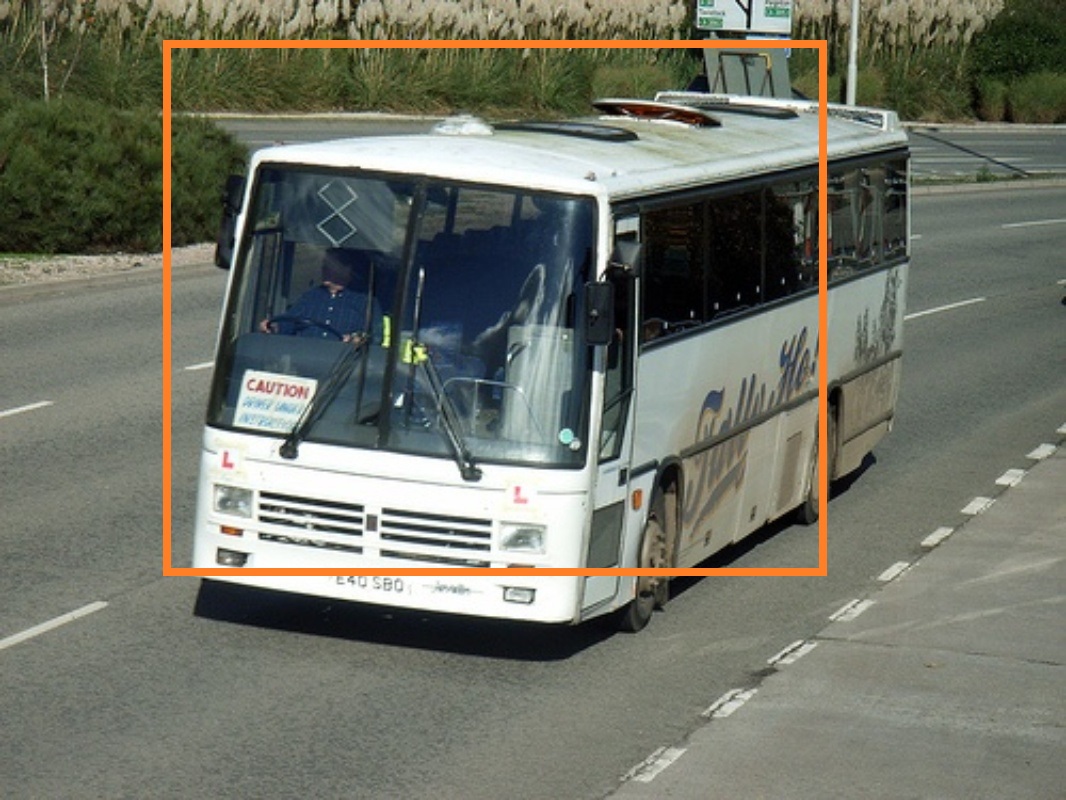}\\ \vspace{-0.1in}
    \includegraphics[width=1.0in,height=1.0in]{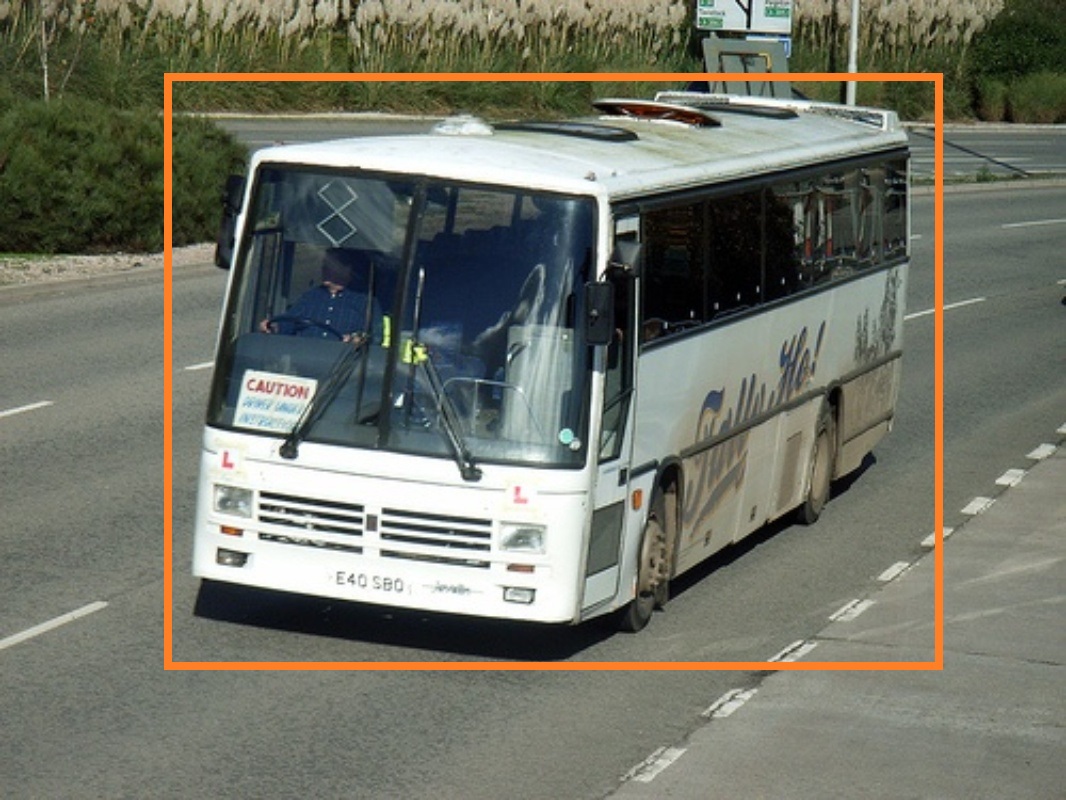}
  \end{minipage}
  }
  \hspace{0.01cm}
  \subfigure{
  \begin{minipage}[t]{0.28\linewidth}
    \includegraphics[width=1.0in,height=1.0in]{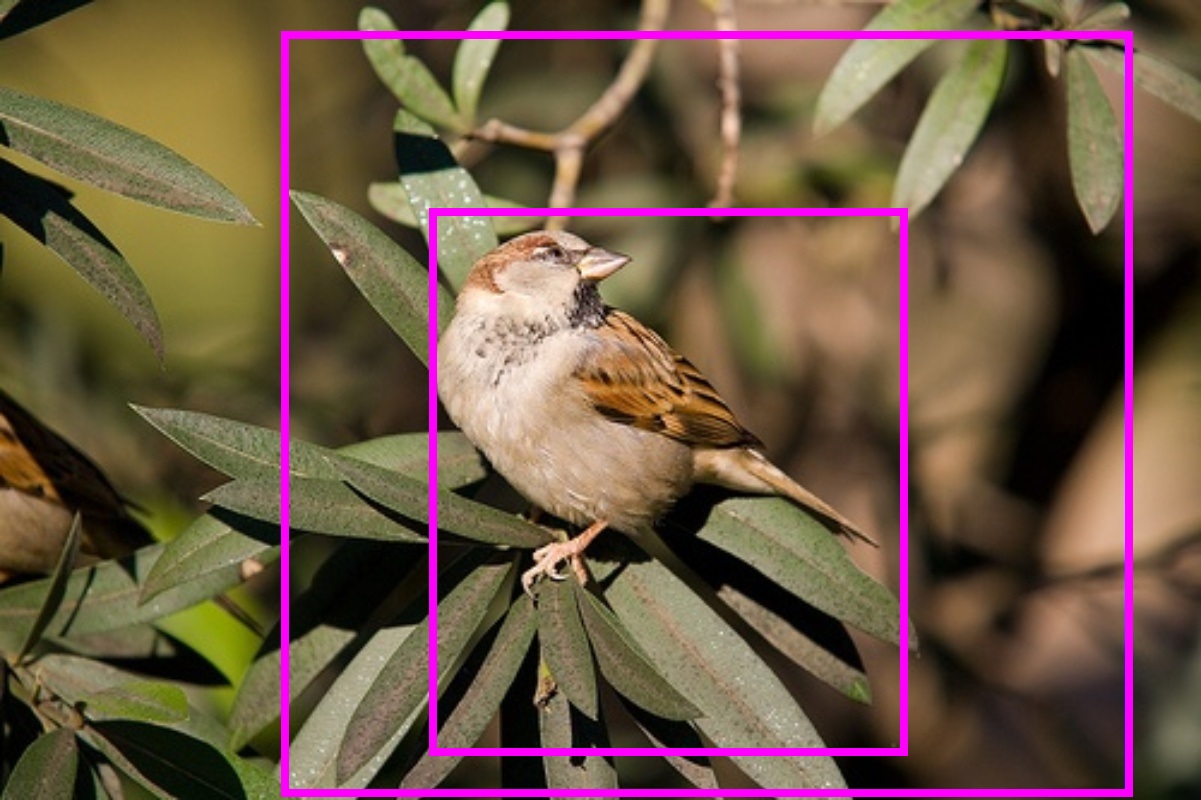}\\ \vspace{-0.1in}
    \includegraphics[width=1.0in,height=1.0in]{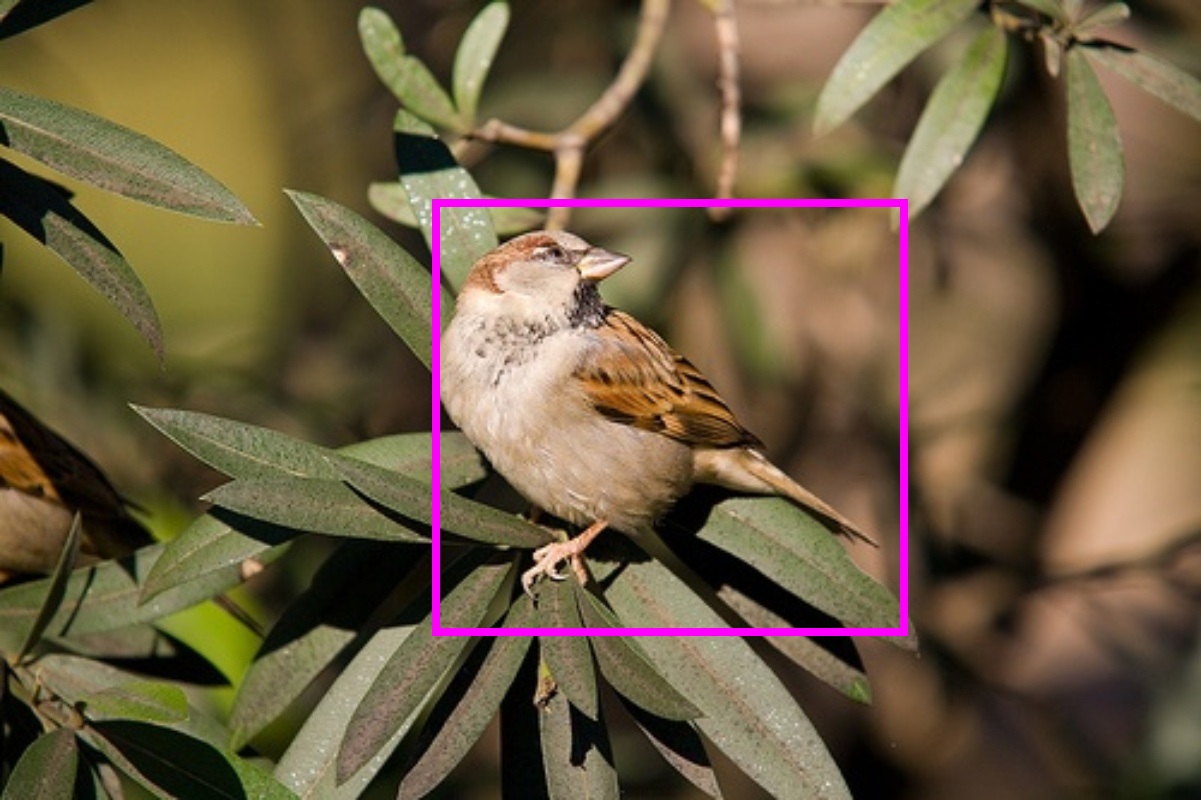}
  \end{minipage}
  }
\end{center}
\vspace{-0.1in}
\caption{Detection results based on the 5-shot case. The first row shows the results of MPSR \cite{wu2020multi}. The second row is our detection results. Our method detects the objects accurately.} \label{fig51}
\vspace{-0.15in}
\end{figure}

In Fig. \ref{fig51}, we show the detection results of MPSR \cite{wu2020multi} and our method. `bird' and `bus' belong to the novel categories. We can see that our method can successfully detect objects existing in images. This further shows that the proposed universal-prototype enhancement is helpful for capturing invariant object characteristics, which improves the accuracy of detection.

\textbf{Results on MS COCO.} Table \ref{table2} shows the few-shot detection performance on MS COCO dataset. Compared with two baseline methods, i.e., TFA \cite{wang2020frustratingly} and MPSR \cite{wu2020multi}, our method consistently outperforms their performance. This further demonstrates the effectiveness of the proposed universal-prototype enhancement. Besides, FSOD-VE \cite{xiao2020few} is a recently proposed meta-learning based method, which combines FSOD with a few-shot viewpoint estimation and follows Meta R-CNN \cite{yan2019meta} to optimize detectors. Though FSOD-VE's performance of the 10-shot case is higher than our method, our method outperforms FSOD-VE on the small objects. Meanwhile, compared with FSOD-VE, the training of our method is much easier. And we do not use the viewpoint information. These results further demonstrate that exploiting universal-prototype enhancement is helpful for improving detectors' generalization.

\begin{table}[t]
\addtolength{\tabcolsep}{2pt}
\begin{center}
\scalebox{0.74}{
\begin{tabular}{c l |ccc cc}
	\toprule
	Shots & Method & AP & AP75 & ${\rm AP}_{\rm S}$ & ${\rm AP}_{\rm M}$ & ${\rm AP}_{\rm L}$ \\
	\midrule
	\multirow{9}{*}{10}
    & Meta R-CNN~\cite{yan2019meta} & 8.7 & 6.6 & 2.3 & 7.7 & 14.0 \\
    & FSOD-VE~\cite{xiao2020few}  & {\bf 12.5} & 9.8 & 2.5 & {\bf 13.8} & {\bf 19.9} \\ \cmidrule(l){2-7}
    & TFA w/fc~\cite{wang2020frustratingly} & 10.0 & 9.2 & -- & -- & -- \\
	& TFA w/cos~\cite{wang2020frustratingly} & 10.0 & 9.3 & -- & -- & -- \\
	& ${\rm TFA}^{\sharp}$ w/fc~\cite{xiao2020few,wang2020frustratingly} & 9.1 & 8.5 & -- & -- & -- \\
	& ${\rm TFA}^{\sharp}$ w/cos~\cite{xiao2020few,wang2020frustratingly} & 9.1 & 8.8 & -- & -- & -- \\
    & ${\rm MPSR}^{\sharp}$~\cite{wu2020multi} & 9.5 & 9.5 & 3.3 & 8.2 & 15.9 \\ \cmidrule(l){2-7}
	& Ours (${FSOD}^{up}$) & 11.0 & {\bf 10.7} & {\bf 4.5} & 11.2 & 17.3 \\
	\midrule
	
	\multirow{9}{*}{30}
    & Meta R-CNN~\cite{yan2019meta} & 12.4 & 10.8 & 2.8 & 11.6 & 19.0 \\
    & FSOD-VE~\cite{xiao2020few}  & 14.7 & 12.2 & 3.2 & {\bf 15.2} & 23.8 \\ \cmidrule(l){2-7}
    & TFA w/fc~\cite{wang2020frustratingly} & 13.4 & 13.2 & -- & -- & -- \\
	& TFA w/cos~\cite{wang2020frustratingly} & 13.7 & 13.4 & -- & -- & -- \\
	& ${\rm TFA}^{\sharp}$ w/fc~\cite{xiao2020few,wang2020frustratingly} & 12.0 & 11.8 & -- & -- & -- \\
	& ${\rm TFA}^{\sharp}$ w/cos~\cite{xiao2020few,wang2020frustratingly} & 12.1 & 12.0 & -- & -- & -- \\
    & ${\rm MPSR}^{\sharp}$~\cite{wu2020multi} & 13.8 & 13.5 & 4.0 & 12.9 & 22.9 \\ \cmidrule(l){2-7}
	& Ours (${FSOD}^{up}$) & {\bf 15.6} & {\bf 15.7} & {\bf 4.7} & 15.1 & {\bf 25.1} \\
	\bottomrule
\end{tabular}}
\vspace{0.05in}
\caption{Few-shot detection performance (\%) on MS COCO dataset. Here, ${\rm AP}_{\rm S}$, ${\rm AP}_{\rm M}$, and ${\rm AP}_{\rm L}$ separately indicate the mAP performance of the small, medium, and large objects.}
\label{table2}
\end{center}
\vspace{-0.15in}
\end{table}

\begin{table}[ht]
  \begin{center}
  \small
  \begin{tabular}{c|ccccc}
  \toprule
  method/shot & 1 & 2 & 3 & 5 & 10 \\ \midrule
  No Condition & 38.1 & 43.8 & 48.9 & 55.6 & 60.6 \\ \midrule
  New Prototype & 42.1 & 44.6 & 48.8 & \textbf{56.1} & 60.1 \\ \midrule
  Ours         & \textbf{43.8} & \textbf{47.8} & \textbf{50.3} & 55.4 & \textbf{61.7} \\
  \bottomrule
  \end{tabular}
  \end{center}
  \vspace{-0.05in}
  \caption{Analysis of conditional prototypes. Here, `No Condition' indicates we do not use conditional operation in Eq. \eqref{eq3} and directly use the universal prototypes $C$ to make enhancement. `New Prototype' indicates we define a new set of prototypes to replace the conditional prototypes.}\label{table3}
  \vspace{-0.15in}
\end{table}

\subsection{Ablation Analysis}

In this section, based on the Novel Set 1 of PASCAL VOC, we make an ablation analysis of our method.

\textbf{Conditional prototypes.} In order to sufficiently represent object-level information, based on the universal prototypes $C$ (see Eq. \eqref{eq1}), we make an affine transformation to obtain conditional prototypes $\mathcal{A}$ (see Eq. \eqref{eq3}). Next, we make an ablation analysis of conditional prototypes.

\begin{figure*}[ht]
  \centering
  \includegraphics[width=1.0\linewidth]{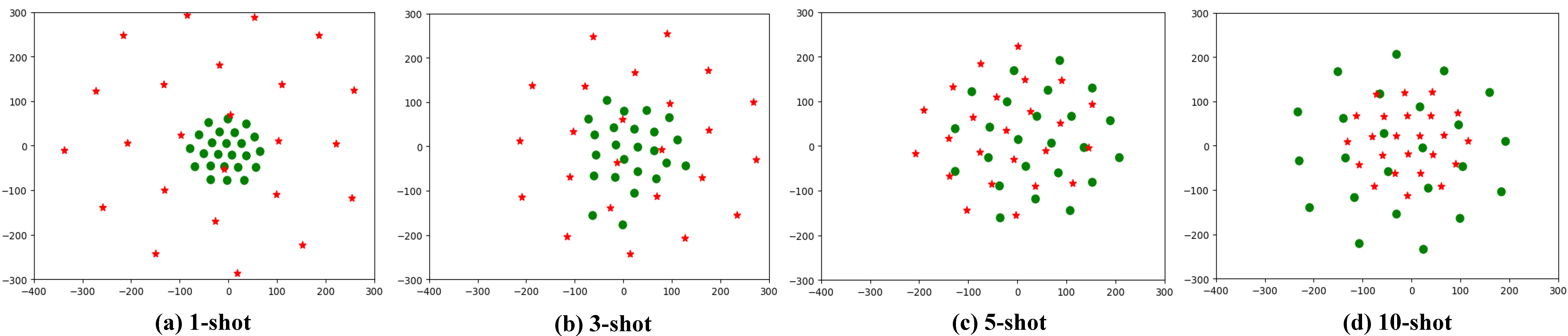}
  \caption{The t-SNE plot of prototypes. We analyze the impact of employing different shots. Here, the number of prototypes is 24. \textcolor[rgb]{0,0.8,0.5}{$\bullet$} and \textcolor[rgb]{1,0,0}{$\bigstar$} separately denote the universal prototypes (see Eq. \eqref{eq1}) and conditional prototypes (see Eq. \eqref{eq3}). For novel categories, using a different number of samples affects the distribution of the universal and conditional prototypes. As the number of novel objects increases, the universal prototypes become more scattered, whereas the conditional ones become more concentrated.} \label{fig5}
  \vspace{-0.05in}
\end{figure*}

\begin{table}[ht]
  \begin{center}
  \small
  \begin{tabular}{c|ccccc}
  \toprule
  number/shot & 1 & 2 & 3 & 5 & 10 \\ \midrule
  16 & 41.2 & 42.7 & 48.3 & 54.2 & 60.1 \\
  20 & 42.5 & 44.1 & 50.1 & 56.0 & 60.5 \\
  24 & \textbf{43.8} & \textbf{47.8} & \textbf{50.3} & 55.4 & \textbf{61.7} \\
  28 & 42.6 & 44.6 & 49.6 & \textbf{56.7} & 60.6 \\
  32 & 41.4 & 42.1 & 49.6 & 53.9 & 60.0 \\
  \bottomrule
  \end{tabular}
  \end{center}
  \vspace{-0.05in}
  \caption{The impact of the number of universal prototypes. Here, we only utilize a different number of prototypes and keep other components unchanged.}\label{table4}
  \vspace{-0.05in}
\end{table}

\begin{table}[ht]
	\begin{center}
     \scalebox{0.7}{
		\begin{tabular}{cc|ccccc|cc}
			\toprule
			&            & \multicolumn{5}{c|}{Novel Classes}                                            & \multicolumn{2}{c}{Mean}     \\ \midrule
			\multicolumn{1}{c|}{Shot}        & Method            & bird          & bus           & cow           & mbike         & sofa          & Novel         & Base          \\ \midrule
			\multicolumn{1}{c|}{\multirow{2}{*}{2}}  & ${\rm MPSR}^{\sharp}$~\cite{wu2020multi}  & 36.8          & 24.8          & 56.9          & 59.1          & 28.4          & 41.2          & 65.4          \\
			\multicolumn{1}{c|}{}  & Ours (${FSOD}^{up}$)  & \textbf{40.7} & \textbf{41.3} & \textbf{58.9} & \textbf{62.2} & \textbf{35.9} & \textbf{47.8} & \textbf{66.3} \\ \midrule
			\multicolumn{1}{c|}{\multirow{2}{*}{5}} & ${\rm MPSR}^{\sharp}$~\cite{wu2020multi}   & 44.1 & \textbf{60.7}          & 54.3          & \textbf{66.8}         & 42.1          & 53.6          & 69.5          \\
			\multicolumn{1}{c|}{}  & Ours (${FSOD}^{up}$)  & \textbf{47.0}          & 60.5 & \textbf{57.3} & 66.4 & \textbf{46.1} & \textbf{55.4} & \textbf{69.7} \\ \bottomrule
		\end{tabular}}
    \vspace{-0.01in}
    \caption{AP (\%) of each novel category on the 2-/5-shot case. We also present mAP (\%) of novel and base categories.} \label{table5}
    \vspace{-0.25in}
	\end{center}
\end{table}

Table \ref{table3} shows the comparison results. We can see that utilizing the conditional operation improves detection performance significantly. Particularly, for the 2-shot case, our method separately outperforms `No Condition' and `New Prototype' by 4.0\% and 3.2\%. This shows that based on the universal prototypes, the conditional prototypes represent object-level information effectively, which improves the performance of detection.

\textbf{The number of universal prototypes.} For our method, the number of universal prototypes (see Eq. \eqref{eq1}) is an important hyper-parameter. If the number is small, these prototypes could not represent invariant object characteristics sufficiently. On the contrary, a large number of prototypes may increase parameters and computational costs.

Table \ref{table4} shows the performance of employing a different number of prototypes. We can see that the performance of utilizing 24 prototypes is the best. When the number is larger or fewer than 24, the performance degrades significantly. This shows the number of prototypes affects FSOD performance. In general, for the case of a large-scale dataset with a large number of categories, employing more prototypes could capture object-level characteristics sufficiently, which is helpful for improving detectors' generalization on novel object categories.

\begin{figure}[ht]
\vspace{-0.1in}
\begin{center}
  \subfigure{
  \begin{minipage}[t]{0.28\linewidth}
    \includegraphics[width=1.0in,height=1.1in]{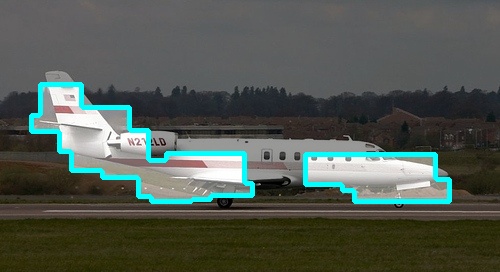} \\ \vspace{-0.1in}
    \includegraphics[width=1.0in, height=1.1in]{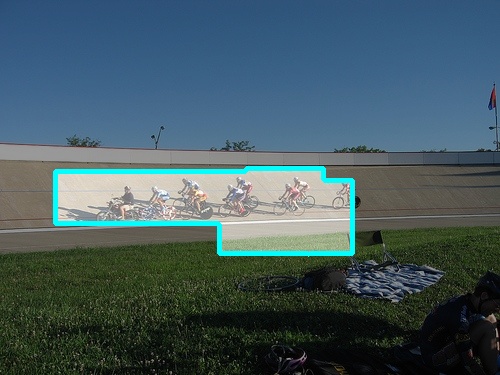}
  \end{minipage}
  }
  \hspace{0.01cm}
  \subfigure{
  \begin{minipage}[t]{0.28\linewidth}
    \includegraphics[width=1.0in, height=1.1in]{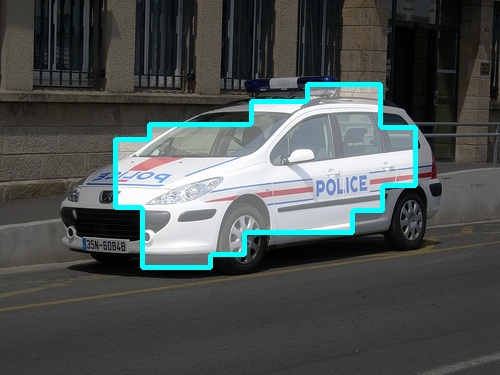}\\ \vspace{-0.1in}
    \includegraphics[width=1.0in,height=1.1in]{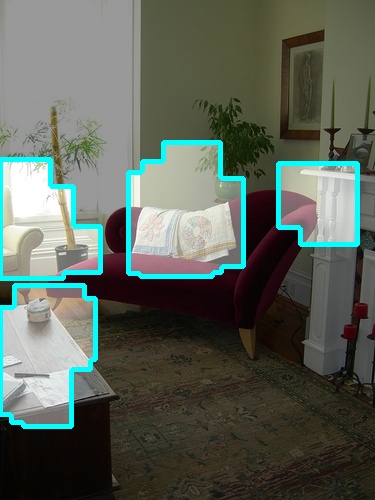}
  \end{minipage}
  }
  \hspace{0.01cm}
  \subfigure{
  \begin{minipage}[t]{0.28\linewidth}
    \includegraphics[width=1.0in,height=1.1in]{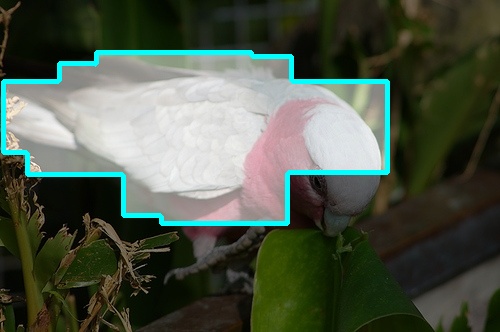}\\ \vspace{-0.1in}
    \includegraphics[width=1.0in,height=1.1in]{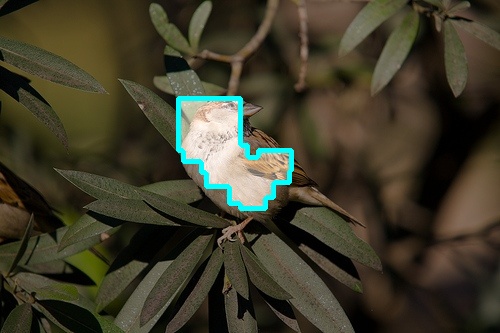}
  \end{minipage}
  }
\end{center}
\vspace{-0.15in}
\caption{Assignment of image regions to universal prototypes based on the 5-shot case. The highlight regions in each image are assigned to one same prototype, respectively.} \label{fig70}
\vspace{-0.25in}
\end{figure}

\textbf{Visualization analysis of prototype distribution.} In Fig. \ref{fig5}, based on different shots, we analyze the distribution of prototypes. Concretely, as the number of novel objects increases, in order to improve the detection performance, the universal prototypes (see Eq. \eqref{eq1}) will become more scattered to capture more image-level information. After RPN, the conditional prototypes are calculated to represent object-level information. And the features calculated based on the conditional prototypes are used for classification. Thus, as the number of novel objects increases, the distribution of the conditional prototypes will become more concentrated to focus on specific categories, which could improve the accuracy of detection. These analyses further show universal prototypes are capable of enhancing feature representations, which is beneficial for FSOD.

\begin{figure*}[ht]
\begin{center}
  \subfigure[\small \textbf{1-shot}]{
  \begin{minipage}[t]{0.18\linewidth}
    \includegraphics[width=1.3in, height=1.00in]{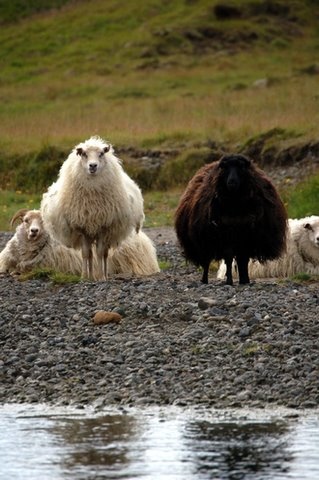}\\ \vspace{-0.1in}
    \includegraphics[width=1.3in, height=1.00in]{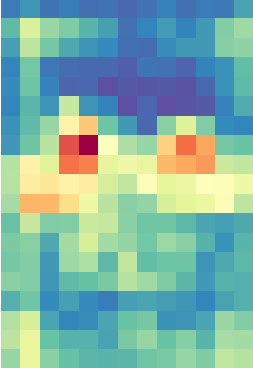}\\ \vspace{-0.1in}
    \includegraphics[width=1.3in, height=1.00in]{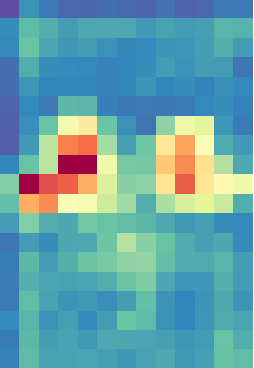}
  \end{minipage}
  }
  \hspace{0.01cm}
  \subfigure[\small \textbf{2-shot}]{
  \begin{minipage}[t]{0.18\linewidth}
    \includegraphics[width=1.3in, height=1.00in]{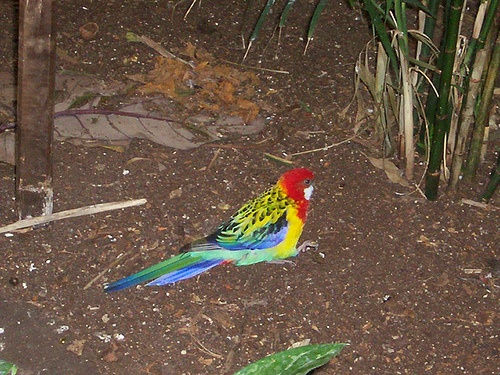}\\ \vspace{-0.1in}
    \includegraphics[width=1.3in, height=1.00in]{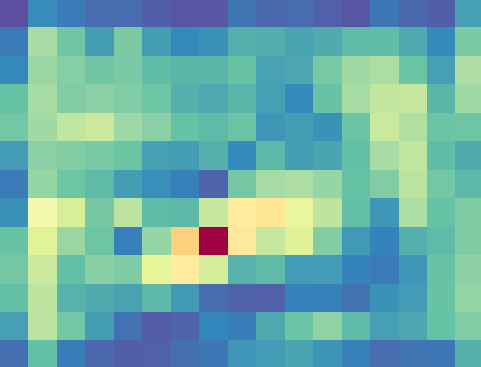}\\ \vspace{-0.1in}
    \includegraphics[width=1.3in, height=1.00in]{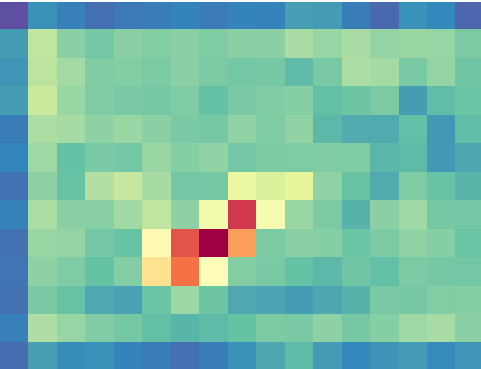}
  \end{minipage}
  }
  \hspace{0.01cm}
  \subfigure[\small \textbf{3-shot}]{
  \begin{minipage}[t]{0.18\linewidth}
    \includegraphics[width=1.3in, height=1.00in]{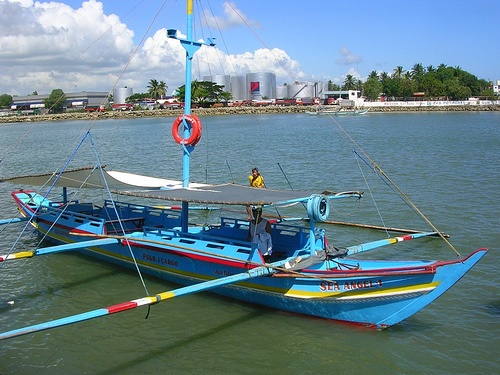}\\ \vspace{-0.1in}
    \includegraphics[width=1.3in, height=1.00in]{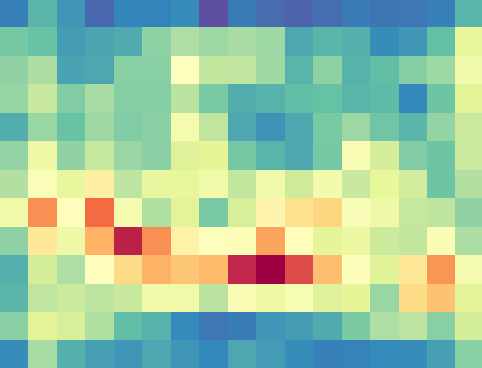}\\ \vspace{-0.1in}
    \includegraphics[width=1.3in, height=1.00in]{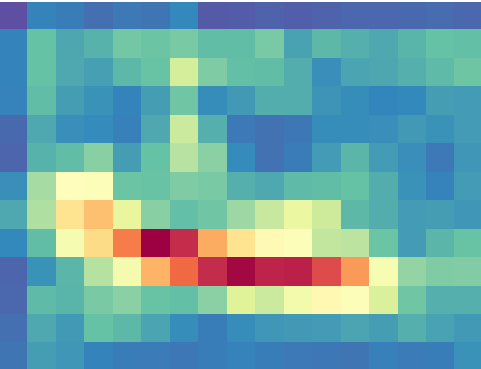}
  \end{minipage}
  }
  \hspace{0.01cm}
  \subfigure[\small \textbf{5-shot}]{
  \begin{minipage}[t]{0.18\linewidth}
    \includegraphics[width=1.3in,height=1.0in]{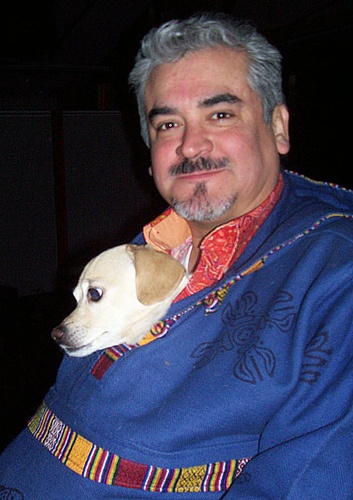}\\ \vspace{-0.1in}
    \includegraphics[width=1.3in,height=1.0in]{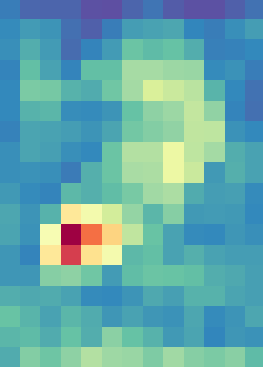}\\ \vspace{-0.1in}
    \includegraphics[width=1.3in,height=1.0in]{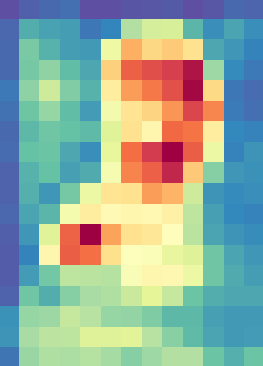}
  \end{minipage}
  }
  \hspace{0.01cm}
  \subfigure[\small \textbf{10-shot}]{
  \begin{minipage}[t]{0.18\linewidth}
    \includegraphics[width=1.3in, height=1.00in]{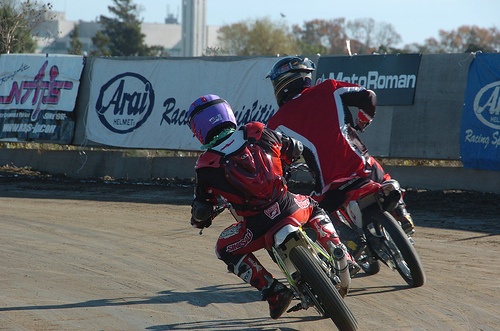}\\ \vspace{-0.1in}
    \includegraphics[width=1.3in, height=1.00in]{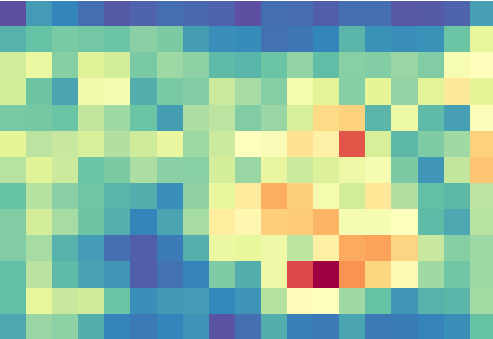}\\ \vspace{-0.1in}
    \includegraphics[width=1.3in, height=1.00in]{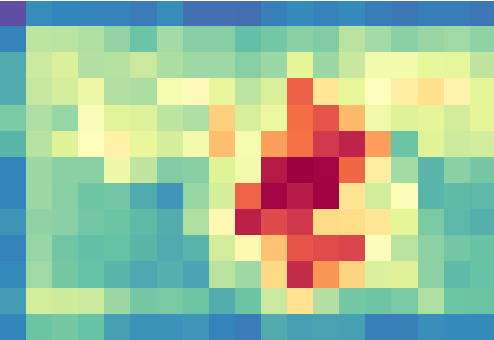}
  \end{minipage}
  }
\end{center}
\vspace{-1em}
\caption{Visualization of the feature map used for RPN based on different shots. The second and third row separately indicate $F$ and the output of $\Psi$ (see Eq. \eqref{eq2}). For each feature map, the channels corresponding to the maximum value are selected for visualization.}
\vspace{-0.1in}
\label{fig80}
\end{figure*}

\textbf{Visualization of assignment maps.} In Fig. \ref{fig70}, we visualize the assignment maps of universal prototypes, i.e., the soft-assignment $\frac{e^{\mathcal{I}_{j,i}}}{\sum_{i=1}^{D} e^{\mathcal{I}_{j,i}}}$ in Eq. \eqref{eq1}. For each image, we can see that different object regions are assigned to one same universal prototype. Particularly, for the second image of the second row, the object regions of `sofa' and `table' are all assigned to one same prototype. This indicates the universal prototypes are not specific to certain object categories. Moreover, the universal prototypes are helpful for characterizing the region information of different objects and could be effectively adapted to novel categories via fine-tuning.
\begin{table}[ht]
  \begin{center}
  \small
  \begin{tabular}{c|ccccc}
  \toprule
  setting/shot & 1 & 2 & 3 & 5 & 10 \\ \midrule
  1.4 & 43.2 & 43.5 & 49.0 & 54.8 & 61.0 \\
  1.2 & 41.1 & 42.1 & 50.2 & 54.3 & 60.7 \\
  1.0 & \textbf{43.8} & \textbf{47.8} & \textbf{50.3} & 55.4 & \textbf{61.7} \\
  0.8 & 39.7 & 42.5 & 49.0 & 56.0 & 60.5 \\
  0.6 & 40.8 & 43.3 & 50.1 & \textbf{56.9} & 60.6 \\
  \bottomrule
  \end{tabular}
  \end{center}
  \vspace{-0.1in}
  \caption{Ablation analysis of the hyper-parameter $\gamma$.}\label{table6}
  \vspace{-0.25in}
\end{table}

\textbf{The performance of base categories.} Table \ref{table5} shows the performance of each novel and base categories. We can see that our method outperforms MPSR \cite{wu2020multi} on novel and base categories. Particularly, for the `bus' and `sofa' category of the 2-shot case, our method outperforms MPSR by 16.5\% and 7.5\%. This indicates our method could improve the generalization performance of the detector.

\textbf{Analysis of Hyper-Parameter $\gamma$.} For the joint training loss $\mathcal{L}$ (see Eq. \eqref{eq8})), we use a hyper-parameter $\gamma$ to balance the consistency loss $\mathcal{L}_{\rm con}$. Table \ref{table6} shows the results. We can see that different settings of the hyper-parameter $\gamma$ affect the performance of FSOD. For our method, when $\gamma$ is set to 1.0, the performance is the best.

\textbf{Analysis of the output descriptors.} In Eq. \eqref{eq2} and \eqref{eq5}, the output descriptors are fused as the input of the RPN and classifier. Next, we analyze the impact of the descriptors. Concretely, for Eq. \eqref{eq2}, we only take $F$ as the input of RPN and keep other components unchanged. For the 1-shot and 5-shot case, fusing the descriptors improves the performance by 2.7\% and 1.8\%. For Eq. \eqref{eq5}, we only take $\Psi_{c}(P)$ as the input of classifier and keep other components unchanged. For the 1-shot and 5-shot case, fusing the descriptors improves the performance by 2.1\% and 1.2\%. This shows fusing descriptors into the current features is helpful for improving the representation ability of the features.

In Fig. \ref{fig80}, based on different shots, we show visualization results of $F$ and the output of $\Psi$ (see Eq. \eqref{eq2}). Here, we separately take $F$ and the output of $\Psi$ as the input of RPN. We can see that for the base and novel categories, compared with $F$, the output of $\Psi$ contains more object-related information. Taking the 5-shot result as an example, the output of our method (the fourth image of the third row) contains more information about `Person' category. This further indicates fusing descriptors is helpful for enhancing the object-level information.

\section{Conclusion}

To solve FSOD, we propose to learn universal prototypes from all object categories. Meanwhile, we develop an approach of few-shot object detection with universal prototypes (${FSOD}^{up}$). Concretely, after obtaining the universal and conditional prototypes, the enhanced object features are computed based on the conditional prototypes. Next, through a consistency loss, ${FSOD}^{up}$ enhances the invariance and generalization. Experimental results on two datasets show the effectiveness of the proposed method.

\section*{Acknowledgement}

This work is supported by the NSFC (under Grant 61876130, 61932009).

{\small
\bibliographystyle{ieee_fullname}
\bibliography{egbib}
}

\end{document}